\documentclass{article}

\usepackage[preprint]{neurips_2026}

\usepackage[utf8]{inputenc} % allow utf-8 input
\usepackage[T1]{fontenc}    % use 8-bit T1 fonts
\usepackage{hyperref}       % hyperlinks
\usepackage{url}            % simple URL typesetting
\usepackage{booktabs}       % professional-quality tables
\usepackage{amsfonts}       % blackboard math symbols
\usepackage{nicefrac}       % compact symbols for 1/2, etc.
\usepackage{microtype}      % microtypography
\usepackage{xcolor}         % colors
\usepackage{amsmath}
\usepackage{algorithm}
\usepackage[noend]{algpseudocode}
\newcommand{\method}{AsyncLane}
\newcommand{\mask}{\texttt{[MASK]}}

\usepackage{multirow}
\usepackage{adjustbox}
\usepackage{graphicx}

\newcommand{\blfootnote}[1]{%
  \begingroup
  \renewcommand\thefootnote{}%
  \renewcommand\thempfn{}%
  \footnotetext{#1}%
  \endgroup
}

\usepackage{booktabs}
\usepackage{makecell}
\usepackage{xcolor}
\usepackage{amssymb}

\newcommand{\cmark}{\textcolor{green!60!black}{\checkmark}}
\newcommand{\xmark}{\textcolor{red!75!black}{$\times$}}

\usepackage{pifont}
\usepackage{xcolor}
\usepackage{makecell}
\usepackage{booktabs}
\usepackage{makecell}
\usepackage{capt-of}
\usepackage{xcolor}
\usepackage{amssymb}
\usepackage{booktabs}
\usepackage{multirow}
\usepackage{graphicx}
\usepackage{footmisc}
\title{AsyncLane: Decoupling Refinement from Advancement in Diffusion Language Model Decoding}

\author{%
\begin{tabular}{cccc}
\multicolumn{4}{c}{Yingxuan Ren$^{1,*}$ \quad Yuxuan Lou$^{1,*}$ \quad Yong Liu$^{1,*}$ \quad Pengcheng Fang$^{2}$} \\
\multicolumn{4}{c}{Ziming Wang$^{1}$ \quad Pengfei Zhou$^{1}$ \quad Yang You$^{1,\dagger}$} \\[3pt]
\multicolumn{4}{c}{$^{1}$National University of Singapore \quad $^{2}$University of Southampton}\\
\end{tabular}
}

\begin{document}

\maketitle
\blfootnote{$^{*}$Equal contribution. \quad $^{\dagger}$Corresponding author. \quad Correspondence to: Yang You <yangyou@nus.edu.sg>.}

\begin{abstract}
Block-wise semi-autoregressive decoding has become the dominant inference paradigm for diffusion large language models, but it imposes a strict serial dependency between blocks: the current block must be fully decoded, or exhaust its denoising budget, before the next block can begin. This serialization places two qualitatively distinct operations on the same critical path: refinement of residual low-confidence positions in the current block, and advancement into the
next block. However, we observe that once the current block has already formed a reliable delimiter boundary or stable semantic prefix, the continuation need not always wait for every remaining token in the current block to be resolved. Building on this, we propose \textbf{AsyncLane}, a training-free decoding scheduler that decouples refinement from advancement in diffusion language model
decoding. AsyncLane forks a generate lane at observed delimiter boundaries into a refine lane and a continuation generate lane: the prefix remains editable, while the continuation starts advancing before prefix refinement finishes. Forking induces a lane tree that records decoding dependencies and output reconstruction order, while execution proceeds over the active lane set. To make this asynchronous schedule efficient under bidirectional attention, AsyncLane uses shared-prefix lane batching, lookahead draft reuse, cascading termination, and compact cache refresh with refresh-logit reuse, keeping model-call cost from scaling directly with the number of lanes. AsyncLane is a drop-in replacement for block-wise DLM samplers and requires no model retraining. Experiments on mathematical reasoning and code generation benchmarks show that
AsyncLane consistently improves throughput while maintaining competitive generation quality. Across both LLaDA and Dream backbones, AsyncLane achieves the highest TPS in all evaluated benchmark-length settings; relative to the fastest competing baseline in each setting, it reaches peak speedups of $2.95\times$ on LLaDA-based setting and $3.04\times$ on Dream-based setting, with especially large gains under longer generation budgets. Our code is available at https://github.com/renyingxuan/AsyncLane.git
\end{abstract}

\section{Introduction}

% Diffusion language models(DLMs) have recently emerged as a promising alternative to autoregressive language models \citep{li2022diffusionlm,lou2023sedd,sahoo2024mdlm,nie2025llada,ye2025dream}. Instead of generating tokens strictly from left to right, they initialize the generation region with mask
% tokens and iteratively denoise the sequence. This paradigm is naturally
% parallelizable: multiple masked positions can be predicted within a single
% forward pass. In practice, however, efficient decoding remains challenging.
% Existing diffusion language models commonly rely on block-wise semi-autoregressive decoding, where the generation region is partitioned into blocks and decoded sequentially ~\citep{sahoo2024mdlm,nie2025llada,wu2025fastdllm}. Although this improves generation stability, it also reintroduces a block-level execution order that limits the parallel potential of diffusion decoding.

% While block-wise decoding improves generation stability and simplifies cache management~\citep{wu2025fastdllm,hu2025flashdlm,ma2025dkvcache}, it also reintroduces a strong serialization constraint. The decoder must keep refining the current block until it is completed or its denoising budget is exhausted before activating the next block. This schedule couples two operations that need not always be synchronized: refinement of the current region and advancement of the generation frontier. Consequently, even when the current block has already exposed a reliable delimiter boundary or a stable prefix, downstream generation may still be forced to wait.

Diffusion language models (DLMs) have recently emerged as a promising alternative to autoregressive language models~\citep{li2022diffusionlm,lou2023sedd,sahoo2024mdlm,nie2025llada,ye2025dream}. Instead of generating tokens strictly from left to right, they initialize the generation region with mask tokens and iteratively denoise the sequence,
allowing multiple masked positions to be predicted in one forward pass. In practice, however, efficient inference commonly relies on block-wise semi-autoregressive decoding, where the generation region is partitioned into blocks and decoded sequentially~\citep{sahoo2024mdlm,nie2025llada,wu2025fastdllm}. While this improves stability and simplifies cache reuse, it imposes a block completion barrier: the decoder must keep refining the current block until it is completed or its denoising budget is exhausted before activating the next block. This schedule couples refinement of the current region with advancement of the generation frontier, forcing downstream generation to wait even when the current block has exposed a reliable delimiter boundary or stable prefix.

\begin{figure}[t]
  \centering
  \includegraphics[width=\linewidth]{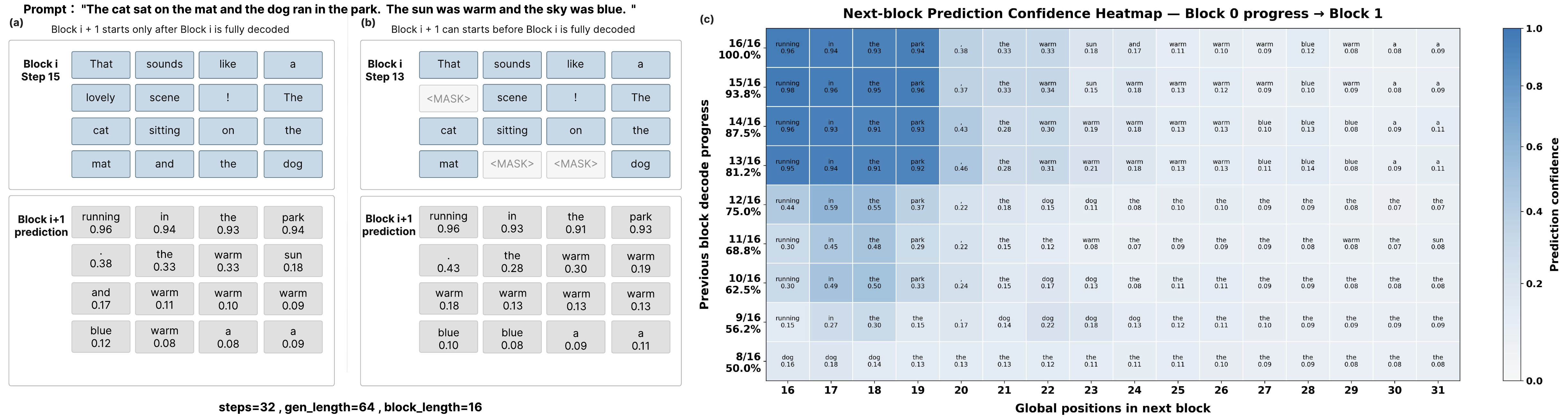}
  \caption{
  \textbf{Motivating phenomenon for asynchronous advancement.}
Even before the current block is fully decoded, the model can assign high confidence to tokens in the next block. \textbf{Panels (a)} and \textbf{(b)} show two intermediate states of block $i$, together with the model's predictions for block $i{+}1$. \textbf{Panel (c)} visualizes next-block prediction confidence as the current block progresses. The observation suggests that waiting for full block completion can be overly conservative, motivating \method{} to start continuation decoding once a reliable boundary is exposed while the prefix remains refinable.
}
\label{fig:motivation}
\end{figure}

This observation motivates our central question: can we decouple refinement from advancement in diffusion language model decoding, without retraining the model? We do not assume that adjacent regions are independent, nor do we aim to merely choose a better block length. Instead, we argue that the dependency between refinement and advancement is not all-or-nothing. Figure~\ref{fig:motivation} illustrates this phenomenon: Once a partial prefix reaches a reliable boundary, the following region can begin denoising while the prefix remains editable. This suggests an asynchronous decoding schedule in which the past continues to be refined while the future starts to advance.

We propose \textbf{AsyncLane}, a training-free decoding scheduler that realizes this idea through active-lane scheduling. AsyncLane replaces the single global
block pointer with a set of active lanes. A generate lane advances the decoding frontier, while a refine lane continues improving a prefix discovered during denoising. When a generate lane exposes a valid delimiter boundary, AsyncLane
performs a \emph{branch-and-refine} operation: the prefix before the boundary is
assigned to a refine lane, and the continuation after the boundary is assigned
to a new generate lane. Both lanes become active, allowing the continuation to
start before prefix refinement finishes. Thus, AsyncLane's asynchrony comes from
changing the decoding dependency structure, not merely from batching multiple
regions for execution.

To make this asynchronous schedule efficient for bidirectional diffusion
language models, AsyncLane exploits the structure created by forking. Forked
sibling lanes share prefix states and can be evaluated with grouped batched
forwards. Future-window logits are reused as high-confidence drafts for later
generate lanes, while cascading termination prevents downstream lanes from being
created beyond EOS or unresolved frontiers. AsyncLane further uses compact cache
refresh and refresh-logit reuse to control stale prefix states under
bidirectional attention. Together, these mechanisms allow AsyncLane to maintain
multiple active lanes without making model-call cost scale directly with the
number of lane nodes.

Experiments on mathematical reasoning and code generation benchmarks show that AsyncLane consistently improves throughput while maintaining competitive generation quality. Across all evaluated benchmark-length settings, AsyncLane achieves the highest throughput among compared decoding methods, and this advantage becomes even more pronounced with a longer generation length.

Our contributions are:
\begin{itemize}
    \item We identify the block completion barrier in block-wise DLM decoding
    and formulate it as a refinement-advancement synchronization problem.
    \item We propose AsyncLane, a training-free active-lane scheduler that
    decouples prefix refinement from frontier advancement through
    boundary-triggered branch-and-refine.
    \item We introduce efficient asynchronous execution mechanisms and show
    empirically, across LLaDA and Dream backbones, that AsyncLane improves
    throughput while maintaining competitive quality.
\end{itemize}

% Our contributions are summarized as follows:
% \begin{itemize}
%     \item We identify the block completion barrier in block-wise diffusion
%     language model decoding and formulate it as a refinement-advancement
%     synchronization problem.

%     \item We propose AsyncLane, a training-free active-lane decoding scheduler
%     that decouples prefix refinement from frontier advancement through
%     boundary-triggered branch-and-refine.

%     \item We introduce efficient asynchronous execution mechanisms, including
%     shared-prefix lane batching, lookahead draft reuse, cascading termination,
%     and compact cache refresh with refresh-logit reuse.

%     \item We empirically show that AsyncLane improves throughput while
%     maintaining competitive quality.
% \end{itemize}

\section{Related Work}

\paragraph{Diffusion language models}
Diffusion-based text generation has been studied as an alternative to
autoregressive language modeling, with early work exploring continuous diffusion
over word embeddings and later work developing discrete or masked diffusion
formulations for language generation
\citep{li2022diffusionlm,lou2023sedd,sahoo2024mdlm}. Recent large-scale
diffusion language models, such as LLaDA and Dream, show that masked denoising
can scale to instruction following, mathematical reasoning, and code generation
\citep{nie2025llada,ye2025dream}. These models offer parallel token refinement
and arbitrary-order generation, but practical inference often still relies on
block-wise semi-autoregressive schedules, which decode blocks sequentially and
therefore reintroduce serialization.

\paragraph{Efficient decoding for DLMs}
Recent work accelerates diffusion language model inference by reducing redundant
computation or improving token-level commit decisions. Fast-dLLM enables
training-free acceleration through approximate KV caching and confidence-aware
parallel decoding~\citep{wu2025fastdllm}. FlashDLM proposes FreeCache and Guided
Diffusion for efficient caching and guided unmasking~\citep{hu2025flashdlm}.
Other cache-based methods, including dKV-Cache, d$^2$Cache, and Elastic-Cache,
study how to reuse or selectively refresh KV states under bidirectional
attention~\citep{ma2025dkvcache,jiang2025d2cache,nguyentri2025elasticcache}.
Complementary approaches such as APD and Learn2PD adapt token-level parallel
decoding decisions~\citep{israel2025apd,bao2025learn2pd}. Adaptive block
scheduling methods such as AdaBlock-dLLM adjust block sizes according to
semantic delimiters and confidence dynamics~\citep{lu2025adablock}.

\method{} is complementary to these methods. Rather than designing a new
token-level commit rule, cache policy, or adaptive block-size heuristic, we
target the temporal scheduling bottleneck: whether advancement to the next
region must wait for refinement of the current region to finish.

\section{Methods}

\begin{figure}[t]
  \centering
  \includegraphics[width=\linewidth]{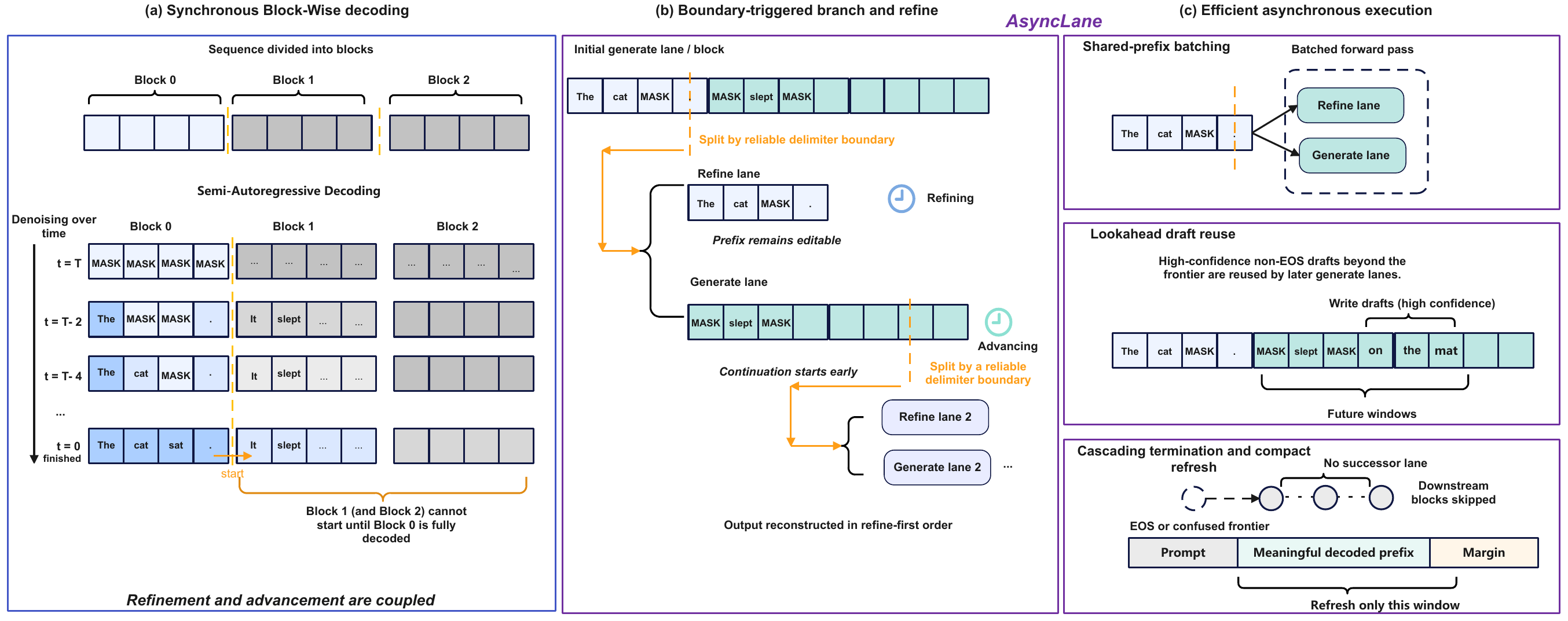}
  \caption{
  \textbf{Overview of \method{}.} 
  \textbf{(a)} Block-wise decoding imposes a synchronous block completion barrier: later
blocks wait until the current block finishes refinement.
\textbf{(b)} \method{} removes this barrier by forking a generate lane at a reliable
delimiter boundary into a refine lane and a continuation generate lane, allowing
the continuation to start before prefix refinement finishes.
\textbf{(c)} Shared-prefix batching, lookahead draft reuse, cascading termination, and compact
cache refresh make the asynchronous schedule efficient.
  }
  \label{fig:asynclane_overview}
\end{figure}

\paragraph{Method overview}
Figure~\ref{fig:asynclane_overview} summarizes \method{}. Standard block-wise decoding advances a single global block pointer, whereas \method{} maintains an active set of lanes and forks generate lanes at reliable delimiter boundaries. We next formalize lane states, branch-and-refine, active-lane scheduling, and efficient execution.
% We present \method{}, a training-free decoding scheduler for bidirectional diffusion language models. Figure~\ref{fig:asynclane_overview} illustrates the full
% decoding procedure. Standard block-wise decoding maintains a single global block pointer, whereas \method{} replaces it with active-lane scheduling. When a generate lane exposes a reliable delimiter boundary, \method{} performs branch-and-refine: the prefix before the boundary is assigned to a refine lane for continued refinement, while the continuation after the boundary is assigned to a new generate lane for early advancement. Repeated branch-and-refine operations induce a lane tree through parent-child relations. This tree records
% decoding dependencies and output reconstruction order, while execution proceeds over the active lane set at each iteration.

% The rest of this section follows this refinement-advancement decoupling view. Section~\ref{sec:method_sync_to_lane} formalizes standard block-wise diffusion decoding and introduces lanes as role-aware scheduling units with independent denoising clocks. Section~\ref{sec:method_branch_refine} describes the
% branch-and-refine procedure built on lane-local denoising and boundary-aware forking. Section~\ref{sec:method_active_lane_scheduling} explains how local forks are composed into an active-lane scheduler and how outputs are reconstructed. Finally, Section~\ref{sec:method_efficiency} presents efficient
% execution mechanisms, including shared-prefix batching, lookahead draft reuse, cascading termination, and compact cache refresh.

\subsection{From Synchronous Block Decoding to Lane-tree Scheduling}
\label{sec:method_sync_to_lane}
% Standard block-wise diffusion decoding is synchronous by construction:
% let $\mathbf{c} = (c_1, \ldots, c_P)$ be the prompt and $L$ the maximum
% generation length. A bidirectional diffusion language model initializes
% the generation region as fully masked and decodes it block by block,
% with the $k$-th block $\mathcal{B}_k$ covering the positions
% $[kB, \min((k+1)B, L))$. 

% At each denoising step, the model proposes tokens for masked positions
% and transfers a subset by confidence; the exact transfer rule varies
% across samplers, but standard block-wise decoding maintains a single
% active block frontier and advances to $\mathcal{B}_{k+1}$ only after
% $\mathcal{B}_k$ has been fully denoised or has exhausted its denoising
% budget. Refinement of the current block and advancement of the
% generation frontier are thereby forced into a lockstep order, which
% can be unnecessarily restrictive: a partially denoised block is often already semantically committed, with its residual low-confidence positions no longer materially shaping the next block, yet the next block must still wait for the current one to fully finish before it can begin.

Standard block-wise diffusion decoding is synchronous by construction. Given a prompt $\mathbf{c}=(c_0,\ldots,c_{P-1})$ of length $P$ and generation length $L$, the model initializes the generation region as fully masked and decodes it block by block, with $\mathcal{B}_k=[kB,\min((k+1)B,L))$. At each denoising step, the sampler transfers a subset of masked positions according to confidence. The key property is that the decoder maintains a single active block frontier: $\mathcal{B}_{k+1}$ is activated only after $\mathcal{B}_k$ is completed or its denoising budget is exhausted.

To break this lockstep dependency, \method{} introduces a \emph{lane} as the
basic scheduling unit. We represent a lane $\ell$ as
\begin{equation}
\ell = (r_\ell, s_\ell, b_\ell, \mathbf{z}_\ell, t_\ell),
\end{equation}

where $r_\ell \in \{\texttt{gen}, \texttt{ref}\}$ is the lane role, $s_\ell$ its start position in generation coordinates, $\mathbf{z}_{\ell}$ is the current token state, $t_{\ell}$ is its local
denoising step. A \emph{generate lane} performs
advancement: it expands the decoding frontier and may reveal future delimiter boundaries.
A \emph{refine lane} performs refinement: it continues improving a prefix that
has already been discovered by a generate lane and does not advance the
frontier. Thus, \method{} replaces the single global block pointer with an active
set of lanes, each progressing according to its own local clock.

\subsection{Asynchronous Branch-and-Refine Decoding}
\label{sec:method_branch_refine}
\method{} starts by defining a single block and performs step-by-step decoding within it. Once a token representing a sentence-level closure is committed inside the block and the block's fill ratio reaches a threshold, \method{} no longer waits for the remaining masked positions in that block to be resolved. Instead, the block is split in place: the prefix up to the sentence end is handed off to a refine lane that continues polishing it, while the positions past the sentence end are handed off to a new generate lane that immediately begins producing the continuation. We call this in-place split at a delimiter boundary \emph{branch-and-refine}.

We describe each iteration through four components: lane-local denoising,
boundary-aware branch point detection, the fork operator, and advancement
control.

\paragraph{Lane-local denoising}
Each active lane is updated independently using its own prefix state. For a lane $\ell$, the prefix state $\kappa_{\ell}$ is the cached KV representation of the context before the lane, up to position $P+s_{\ell}$, where $P$ is the prompt length and $s_{\ell}$ is the lane start position in generation coordinates. The lane input consists of the current lane state followed by an optional lookahead window:
\begin{equation}
    \mathbf{u}_{\ell}
    =
    \left[
    \mathbf{z}_{\ell};
    \mathbf{x}_{P+s_{\ell}+b_{\ell}:P+s_{\ell}+b_{\ell}+w_{\ell}}
    \right],
    \label{eq:lane_input}
\end{equation}
where $b_{\ell}$ is the lane length and $w_{\ell}$ is the lookahead window size. The model is evaluated as
\begin{equation}
    \mathbf{o}_{\ell}
    =
    f_{\theta}(\mathbf{u}_{\ell};\kappa_{\ell}).
\end{equation}

The first $b_{\ell}$ logits of $\mathbf{o}_{\ell}$ are used to update the current
lane with the same confidence-based token transfer rule as the underlying block-wise sampler. 
% In our implementation, this rule follows the Fast-dLLM/LLaDA sampler: masked positions whose prediction confidence exceeds a threshold are
% transferred, and the highest-confidence masked position is transferred when no
% position passes the threshold, ensuring progress. The remaining logits, if a
% lookahead window is present, are reserved for lookahead prefilling.

The only lane-specific modification is the valid update region. For generate
lanes, the entire lane is eligible for transfer. For refine lanes, updates are
restricted to the discovered prefix region, and EOS predictions are suppressed
because refinement should not terminate generation. Generate lanes allow EOS so
that the advancing frontier can stop naturally. After transfer, the updated
tokens are written back to both the lane state $\mathbf{z}_{\ell}$ and the
corresponding global positions in $\mathbf{x}$.

\paragraph{Boundary-aware branch point}
\method{} forks a generate lane only at a reliable structural boundary. Let $\mathcal{D}_{\mathrm{task}}$
denote the task-specific delimiter set. In our main experiments, the exact delimiter set is specified in ~\ref{sec:experimental_setup}; for example, mathematical reasoning tasks use period and Chinese-period delimiters with numeric-period exclusions, while code generation uses newline delimiters.

We take the last observed valid delimiter as the split point,
\begin{equation}
    p_{\ell}
    =
    1+\max\{i:z_{\ell,i}\in\mathcal{D}_{\mathrm{task}}
    \ \text{and}\ 
    \mathrm{ValidBoundary}(z_{\ell,1:i})\}.
    \label{eq:split_point}
\end{equation}

If no valid delimiter is observed, the lane is not forked. To avoid premature
branching, we require sufficient local evidence:
\begin{equation}
    \frac{|\{i:z_{\ell,i}\neq\mask\}|}{b_{\ell}}
    \geq
    \rho_{\min},
    \qquad
    t_{\ell}\geq t_{\min}.
    \label{eq:fork_condition}
\end{equation}

The predicate $\mathrm{ValidBoundary}$ filters delimiters that are unlikely to
mark a stable closure point. For code, this includes avoiding splits inside
unclosed brackets, incomplete statements, or task-specific no-fork patterns.

\paragraph{Fork operator}
Given a generate lane $\ell$ and a valid split point $p_{\ell}$, \method{}
forks it into a refine lane and a continuation generate lane:
\begin{equation}
    \mathcal{F}(\ell,p_{\ell})
    =
    \left(
    \ell^{\mathrm{ref}},
    \ell^{\mathrm{gen}}
    \right).
    \label{eq:fork_operator}
\end{equation}
The refine lane covers the prefix before the boundary, while the continuation
generate lane starts immediately after the boundary and inherits the current
global token state, including any already denoised suffix tokens or lookahead
drafts. Both children become active, so refinement and advancement proceed with
separate lane clocks:
\begin{equation}
    \mathrm{Start}(\ell^{\mathrm{gen}})
    <
    \mathrm{Finish}(\ell^{\mathrm{ref}}).
    \label{eq:branch_async_property}
\end{equation}
This inequality captures the asynchronous property of \method{}: the
continuation no longer waits for prefix refinement to finish. The continuation
uses the boundary prefix state available at fork time, and stale states are
controlled by the cache refresh mechanism in
Section~\ref{sec:method_efficiency}.

\paragraph{Advancement control}
\method{} uses two lightweight rules to control how far advancement proceeds,
as illustrated in Figure~\ref{fig:asynclane_overview}. First, let $\mathcal{E} = {\mathrm{EOS}}$, when a generate-lane forward includes a lookahead window, high-confidence non-EOS
predictions in the future window are written into the global sequence as drafts:
\begin{equation}
    x_j \leftarrow \hat{x}_j
    \quad
    \text{if}
    \quad
    x_j=\mask,\;
    q_j\geq\tau_{\mathrm{future}},\;
    \hat{x}_j\notin\mathcal{E}.
\end{equation}
These drafts are inherited by later generate lanes. Second, after a generate lane completes, \method{} spawns a successor only if the completed lane provides
a non-empty valid prefix. Let
\begin{equation}
    v_{\ell}
    =
    \min
    \left(
    \{i-1: z_{\ell,i}=\mask \ \text{or}\ z_{\ell,i}\in\mathcal{E}\}
    \cup
    \{b_{\ell}\}
    \right).
\end{equation}
If $v_{\ell}=0$, advancement stops from this lane. This cascading rule prevents
downstream lanes from being created beyond EOS or unresolved frontiers. We
analyze the efficiency effect of lookahead reuse and cascading termination in
Section~\ref{sec:method_efficiency}.

\subsection{Active-Lane Scheduling and Output Collection}
\label{sec:method_active_lane_scheduling}

The branch-and-refine operator defines a local transition. To compose these
transitions into a complete decoder, \method{} maintains an active set of
unfinished lanes and records their parent-child relations in
$\mathcal{T}_{\mathrm{lane}}$. The tree records dependencies and output order,
while execution proceeds over the active frontier
\begin{equation}
    \mathcal{A}_t=\{\ell\in\mathcal{T}_{\mathrm{lane}}:\ell\text{ is unfinished}\}.
\end{equation}
At each iteration, the scheduler updates active lanes, applies boundary-aware
forks, and either spawns successor generate lanes or stops advancement from
invalid frontiers. Algorithm~\ref{alg:asynclane} summarizes the procedure.

\begin{algorithm}[t]
\caption{\method{} active-lane decoding}
\label{alg:asynclane}
\begin{algorithmic}[1]
\Require Prompt $\mathbf{c}$, generation length $L$, block size $B$
\State Initialize $\mathbf{x}\gets[\mathbf{c};\mask^L]$
\State Create root generate lane $\ell_0$ and initialize $\mathcal{T}_{\mathrm{lane}}$
\While{there exists an active lane and decoding budget remains}
    \State $\mathcal{A}_t \gets \{\ell\in\mathcal{T}_{\mathrm{lane}}:\ell \text{ is unfinished}\}$
    \State \Call{UpdateLanes}{$\mathcal{A}_t,\mathbf{x}$}
    \For{each generate lane $\ell\in\mathcal{A}_t$}
        \If{$\ell$ satisfies the boundary-aware fork condition}
            \State $(\ell^{\mathrm{ref}},\ell^{\mathrm{gen}})\gets\mathcal{F}(\ell,p_{\ell})$
            \State Attach $\ell^{\mathrm{ref}}$ and $\ell^{\mathrm{gen}}$ to $\mathcal{T}_{\mathrm{lane}}$
            \State Mark $\ell$ as completed
        \ElsIf{$\ell$ is completed}
            \State $v_{\ell}\gets \Call{ValidPrefixLength}{\ell}$
            \If{$v_{\ell}=0$ or the next start position exceeds $L$}
                \State \Call{StopAdvance}{$\ell$}
            \Else
                \State Finalize the valid prefix of $\ell$
                \State Spawn a successor generate lane starting at $s_\ell+v_\ell$ and attach it to $\mathcal{T}_{\mathrm{lane}}$
            \EndIf
        \EndIf
    \EndFor
\EndWhile
\State \Return \Call{CollectDFS}{$\mathcal{T}_{\mathrm{lane}}$}
\end{algorithmic}
\end{algorithm}

\paragraph{Output reconstruction}
After decoding, \method{} traverses the induced lane tree to reconstruct the
sequence. Refine children are emitted before generate children, ensuring that
refined prefixes precede their continuations; leaf lanes are truncated at the
first mask token, and generate lanes are additionally truncated at EOS. This is
a linear post-processing step with no additional model calls.

\subsection{Efficient Asynchronous Execution}
\label{sec:method_efficiency}

A natural concern is that maintaining multiple active lanes may increase the
number of model calls. \method{} avoids this by exploiting the structure created
by forking: sibling lanes share prefix states, future logits are reused by later
lanes, terminated frontiers stop spawning downstream lanes, and cache refresh is
restricted to the meaningful decoded prefix.

\paragraph{Shared-prefix lane batching}
When a generate lane is forked, its refine child and continuation generate child
are created from the same boundary state and therefore share the same fork
prefix state. At iteration $t$, \method{} groups active lanes by shared prefix
state into $\mathcal{G}_t$. Each group is evaluated with one batched model call,
so the model-call cost of the iteration is
\begin{equation}
    \mathrm{NFE}^{\mathrm{call}}_t
    =
    |\mathcal{G}_t|
    \leq
    |\mathcal{A}_t|.
\end{equation}
This does not eliminate token-level computation, but replaces serial lane
forwards with grouped GPU execution. We therefore report both model-call NFE and
wall-clock latency.

\paragraph{Future computation reuse}
The advancement-control rules in Section~\ref{sec:method_branch_refine} also
reduce future computation. Lookahead drafts are inherited by later generate
lanes, so future lanes often start with fewer masked positions. Cascading
termination prevents successor lanes from being spawned when the frontier begins
with EOS or an unresolved mask, short-circuiting downstream computation once the
meaningful frontier stops.

\paragraph{Compact cache refresh and reuse}
Because bidirectional attention makes prefix states stale as earlier tokens are updated, \method{} periodically refreshes them. Instead of refreshing the full prompt-plus-generation sequence, it refreshes only the meaningful prefix plus a
small margin:
\begin{equation}
    L_{\mathrm{refresh}}
    =
    \min(P+L,\; P+L_{\mathrm{eff}}+m),
    \qquad
    L_{\mathrm{eff}}
    =
    \max_{\ell\in\mathcal{T}_{\mathrm{lane}}}(s_\ell+v_\ell).
\end{equation}
The refresh logits are reused in the next iteration for lane-local transfer and lookahead prefilling, amortizing the refresh cost.

\paragraph{Model-call NFE accounting}
Standard block-wise decoding requires roughly
$\mathrm{NFE}_{\mathrm{block}}=N\cdot S$ model calls for
$N=\lceil L/B\rceil$ blocks and $S$ denoising steps per block. In contrast,
\method{} has
\begin{equation}
    \mathrm{NFE}_{\method}
    =
    1+\sum_{t=1}^{T}|\mathcal{G}_t|
    +N_{\mathrm{refresh}}-N_{\mathrm{reuse}}.
\end{equation}
Thus, the cost scales with shared-prefix groups and the meaningful frontier
length, rather than directly with the number of lane nodes.

\section{Experiments}
\subsection{Experimental Setup}
\label{sec:experimental_setup}

% \paragraph{Benchmark Datasets.}
We evaluate \method{} on five representative tasks. For mathematical reasoning, we evaluate on GSM8K~\citep{cobbe2021gsm8k},
MATH~\citep{hendrycks2021math}, and GSM8K-CoT (chain-of-thought reasoning). For code generation, we use
HumanEval~\citep{chen2021humaneval} and MBPP~\citep{austin2021mbpp}. For each backbone, all experiments use the same pretrained masked diffusion language model without fine-tuning. We evaluate decoding efficiency using tokens per second (TPS), and generation quality using exact-match accuracy or pass@1, depending on the benchmark.

To validate the effectiveness of our approach, we compare \method{} framework with state-of-the-art dLLM methods, including vanilla LLaDA~\citep{nie2025llada} and Dream~\citep{ye2025dream}, and the training-free inference acceleration method Fast-dLLM~\citep{wu2025fastdllm}. We additionally evaluate an inference-only variant of d3LLM~\citep{qian2026d3llm}, denoted as d3LLM-TF. Since the full d3LLM framework includes pseudo-trajectory distillation during training, we exclude this training component and use only its inference-side multi-block decoding and cache-refresh strategy on the same pretrained base model.

% \paragraph{Implementation Details.}
For boundary-aware forking, we instantiate the delimiter set according to the task. For mathematical reasoning tasks, the default delimiter set contains period and Chinese-period tokens, with numeric-period patterns excluded. For code generation tasks, we use newline tokens. Unless otherwise specified, \method{} uses the same block size, generation length, and denoising budget as the block-wise baseline. We use a single H100 GPU and fix the batch size to 1 for all models for inference.

\subsection{Evaluation of \method{} Framework}

\paragraph{Results on LLaDA-based Models}
As shown in Table~\ref{tab:main_results_LLaDA}, across all benchmarks and generation budgets, \method{} achieves the highest throughput. This shows that active-lane scheduling does not introduce prohibitive overhead despite maintaining multiple lanes. Instead, the combination of shared-prefix batching, lookahead reuse, cascading termination, and compact cache refresh makes asynchronous decoding substantially more
efficient.

The throughput advantage becomes more pronounced as the generation length increases. Averaged over all benchmarks, the speedup of \method{} over the fastest competing baseline increases from approximately $1.45\times$ at length 256 to $1.62\times$ at length 512 and $2.25\times$ at length 1024. This trend is consistent with the design of \method{}: longer generation budgets create more opportunities for early continuation, future-logit reuse, and downstream
termination, which amplify the benefit of decoupling refinement from advancement.

In terms of generation quality, \method{} maintains competitive accuracy across reasoning and code-generation tasks. On MATH, \method{} achieves the best
accuracy at lengths 512 and 1024 while also delivering the highest TPS. On code benchmarks, \method{} substantially improves throughput while remaining
competitive with training-free acceleration baselines. Overall, the results indicate that \method{} provides a favorable speed-quality tradeoff: it consistently improves practical decoding efficiency while avoiding the large quality degradation that would result from overly aggressive early commitment.

\begin{table}[t]
\centering
\caption{
Comparison of \method{}-LLaDA with other LLaDA based models. We report accuracy (Acc) and throughput in tokens per second (TPS). For GSM8K and MATH, Acc is exact-match accuracy; for MBPP and HumanEval, Acc is pass@1. The best results are highlighted in \textbf{bold}.
}
\label{tab:main_results_LLaDA}
\scriptsize
\setlength{\tabcolsep}{2.6pt}
\renewcommand{\arraystretch}{1.08}
\resizebox{\linewidth}{!}{
\begin{tabular}{@{}l l ccc ccc ccc ccc ccc@{}}
\toprule
\multirow{2}{*}{Method}
& \multirow{2}{*}{Metric}
& \multicolumn{3}{c}{GSM8K-CoT (0-shot)}
& \multicolumn{3}{c}{GSM8K (5-shot)}
& \multicolumn{3}{c}{MATH (4-shot)}
& \multicolumn{3}{c}{MBPP (3-shot)}
& \multicolumn{3}{c}{HumanEval (0-shot)} \\
\cmidrule(lr){3-5}
\cmidrule(lr){6-8}
\cmidrule(lr){9-11}
\cmidrule(lr){12-14}
\cmidrule(lr){15-17}
& & 256 & 512 & 1024
& 256 & 512 & 1024
& 256 & 512 & 1024
& 256 & 512 & 1024
& 256 & 512 & 1024 \\
\midrule

\multirow{2}{*}{LLaDA}
& Acc(\%)$\uparrow$ 
& 73.00 & 74.30 & \textbf{75.13}
& 77.86 & \textbf{79.00} & \textbf{78.85}
& \textbf{32.94} & 31.74 & 31.19
& 40.20 & 39.00 & \textbf{39.40}
& 37.19 & 35.98 & 36.01 \\
& TPS$\uparrow$
& 41.61 & 19.54 & 6.19
& 15.19 & 5.89 & 1.86
& 20.13 & 10.01 & 4.28
& 15.67 & 10.76 & 2.85
& 13.52 & 6.04 & 1.19 \\
\midrule

\multirow{2}{*}{Fast-dLLM-LLaDA}
& Acc(\%)$\uparrow$
& 74.30 & 74.83 & 73.46
& 76.57 & 78.00 & 77.41
& 30.96 & 31.40 & 30.00
& 37.80 & 38.60 & 35.80
& 37.80 & 37.46 & 37.37 \\
& TPS$\uparrow$
& 104.99 & 100.83 & 74.65
& 84.01 & 64.55 & 35.20
& 69.31 & 72.73 & 68.69
& 76.69 & 77.38 & 53.09
& 60.84 & 35.88 & 11.28 \\
\midrule

\multirow{2}{*}{d3LLM-TF-LLaDA}
& Acc(\%)$\uparrow$
& 72.10 & \textbf{74.91} & 74.98
& \textbf{78.24} & 78.24 & \textbf{78.85}
& 31.40 & 31.80 & 31.50
& \textbf{40.80} & 39.60 & 36.00
& \textbf{39.02} & \textbf{44.51} & \textbf{40.24} \\
& TPS$\uparrow$
& 128.82 & 115.74 & 76.80
& 78.85 & 68.33 & 46.21
& 75.46 & 73.56 & 66.02
& 84.37 & 76.20 & 52.93
& 112.61 & 92.52 & 71.72 \\
\midrule

\multirow{2}{*}{\textbf{\method{}-LLaDA}}
& Acc(\%)$\uparrow$
& \textbf{74.91} & 73.77 & 73.39
& 76.42 & 77.33 & 76.80
& 32.52 & \textbf{32.36} & \textbf{32.04}
& 38.40 & \textbf{39.80} & 37.80
& 38.47 & 37.80 & 38.22 \\
& TPS$\uparrow$
& \textbf{166.15} & \textbf{163.30} & \textbf{165.42}
& \textbf{130.29} & \textbf{131.66} & \textbf{136.34}
& \textbf{136.05} & \textbf{138.35} & \textbf{145.40}
& \textbf{112.43} & \textbf{106.26} & \textbf{111.89}
& \textbf{140.58} & \textbf{137.88} & \textbf{138.58} \\
\bottomrule
\end{tabular}
}
\end{table}

\paragraph{Results on Dream-based Models} 
Table~\ref{tab:main_results_Dream} evaluates \method{} on the Dream backbone. Across all benchmarks and generation lengths, \method{}-Dream achieves the highest TPS among the compared Dream-based methods, showing that the proposed
scheduler is not specific to LLaDA. The advantage becomes stronger for longer generation budgets: the average speedup over the fastest non-\method{} baseline increases from about $1.16\times$ at length 256 to $2.49\times$ at length 1024.
This supports our intuition that longer sequences expose more opportunities for early continuation, lookahead reuse, and cascading termination. Accuracy remains competitive overall, although vanilla Dream often has the strongest exact-match
accuracy on reasoning tasks. These results confirm that AsyncLane provides a backbone-agnostic throughput improvement with a reasonable speed-quality tradeoff.

\begin{table}[t]
\centering
\caption{
Comparison of \method{}-Dream with other Dream-based models. We report accuracy
(Acc) and throughput in tokens per second (TPS). For GSM8K and MATH, Acc is
exact-match accuracy; for MBPP and HumanEval, Acc is pass@1. The best results
are highlighted in \textbf{bold}.
}
\label{tab:main_results_Dream}
\scriptsize
\setlength{\tabcolsep}{2.6pt}
\renewcommand{\arraystretch}{1.08}
\resizebox{\linewidth}{!}{
\begin{tabular}{@{}l l ccc ccc ccc ccc ccc@{}}
\toprule
\multirow{2}{*}{Method}
& \multirow{2}{*}{Metric}
& \multicolumn{3}{c}{GSM8K-CoT (0-shot)}
& \multicolumn{3}{c}{GSM8K (5-shot)}
& \multicolumn{3}{c}{MATH (4-shot)}
& \multicolumn{3}{c}{MBPP (3-shot)}
& \multicolumn{3}{c}{HumanEval (0-shot)} \\
\cmidrule(lr){3-5}
\cmidrule(lr){6-8}
\cmidrule(lr){9-11}
\cmidrule(lr){12-14}
\cmidrule(lr){15-17}
& & 256 & 512 & 1024
& 256 & 512 & 1024
& 256 & 512 & 1024
& 256 & 512 & 1024
& 256 & 512 & 1024 \\
\midrule

\multirow{2}{*}{Dream}
& Acc(\%)$\uparrow$
& \textbf{83.17} & \textbf{82.95} & \textbf{82.16}
& \textbf{78.77} & \textbf{78.43} & \textbf{77.83}
& \textbf{38.27} & 38.58 & \textbf{38.29}
& \textbf{60.00} & \textbf{59.40} & \textbf{58.10}
& 47.56 & 47.79 & 47.65 \\
& TPS$\uparrow$
& 16.34 & 4.73 & 1.08
& 11.60 & 4.64 & 1.05
& 17.73 & 7.73 & 4.75
& 5.48 & 2.14 & 0.95
& 9.70 & 4.24 & 1.70 \\
\midrule

\multirow{2}{*}{Fast-dLLM-Dream}
& Acc(\%)$\uparrow$
& 78.09 & 78.32 & 77.52
& 76.95 & 75.97 & 76.42
& 37.98 & 37.59 & 37.70
& 56.40 & 54.80 & 54.40
& \textbf{56.70} & \textbf{54.87} & \textbf{56.70} \\
& TPS$\uparrow$
& 72.50 & 56.13 & 34.60
& 81.10 & 57.15 & 32.34
& 87.06 & 75.50 & 64.62
& 23.86 & 16.99 & 9.79
& 68.25 & 54.30 & 31.28 \\
\midrule

\multirow{2}{*}{d3LLM-TF-Dream}
& Acc(\%)$\uparrow$
& 80.52 & 82.03 & 81.05
& 77.41 & 76.72 & 77.26
& 37.82 & 37.68 & 37.02
& 48.20 & 49.20 & 48.20
& 50.60 & 53.04 & 51.82 \\
& TPS$\uparrow$
& 121.96 & 81.66 & 49.94
& 138.80 & 90.54 & 56.89
& 130.00 & 102.05 & 74.37
& 68.31 & 68.18 & 45.81
& 105.69 & 79.51 & 51.66 \\
\midrule

\multirow{2}{*}{\textbf{\method{}-Dream}}
& Acc(\%)$\uparrow$
& 80.03 & 80.30 & 80.44
& 77.19 & 77.58 & 77.11
& 37.69 & \textbf{38.80} & 37.66
& 54.40 & 54.60 & 54.30
& 53.05 & 51.83 & 52.44 \\
& TPS$\uparrow$
& \textbf{148.60} & \textbf{151.29} & \textbf{151.64}
& \textbf{146.40} & \textbf{147.69} & \textbf{147.71}
& \textbf{141.36} & \textbf{140.83} & \textbf{140.43}
& \textbf{89.69} & \textbf{89.76} & \textbf{90.44}
& \textbf{126.00} & \textbf{115.03} & \textbf{121.72} \\
\bottomrule
\end{tabular}
}
\end{table}

\subsection{Ablation Study}
\paragraph{Ablation on Refinement-Advancement Decoupling}

Table~\ref{tab:decoupling_ablation} isolates the two key components of branch-and-refine. Boundary-aware Coupled uses the same delimiter detector as \method{}, but waits for prefix refinement to finish before starting the continuation. Commit-and-Advance starts the continuation immediately after a detected boundary, but commits the prefix without spawning a refine lane. Full \method{} combines both properties: the prefix remains refinable while the continuation starts early.

The results show that both properties are necessary. Boundary-aware Coupled is more efficient than Block-wise decoding, but remains slower than \method{} because advancement is still gated by refinement completion. Commit-and-Advance achieves low NFE but suffers a large accuracy drop, indicating that prematurely
committing the prefix hurts reasoning quality. \method{} achieves the best speed-quality tradeoff, improving accuracy to 74.91\%, reducing NFE to 60,378, and achieving the highest TPS of 166.15.

Table~\ref{tab:async_diagnostics} confirms that this gain is associated with actual asynchronous execution. Boundary-aware Coupled has zero positive lead and zero overlap by construction, whereas \method{} obtains non-zero advancement lead and overlap. This indicates that continuation lanes measurably start before
their corresponding refine lanes finish, supporting our claim that \method{} decouples refinement from advancement rather than merely changing the boundary detector.

\paragraph{Ablation on Efficient Execution Mechanisms}

Table~\ref{tab:efficiency_ablation} ablates lookahead prefilling and refresh-logit reuse. Since these variants can emit different numbers of tokens, we report NFE/token in addition to TPS. Disabling lookahead causes a large accuracy drop and much shorter outputs; although its raw NFE is lower, its NFE/token increases from 0.288 to 0.399 and TPS drops from 166.15 to 117.8.
This shows that lookahead prefilling helps future generate lanes inherit useful drafts rather than simply reducing computation by early termination. Disabling refresh-logit reuse preserves accuracy and output length, but increases NFE/token and lowers TPS, confirming that refresh logits amortize the cost of cache refresh. The full \method{} achieves the best quality-throughput tradeoff.

\begin{table}[t]
\centering
\scriptsize
\renewcommand{\arraystretch}{1.08}

\begin{minipage}[t]{0.58\linewidth}
\centering
\caption{Ablation study on refinement-advancement decoupling. We report accuracy, NFE, and TPS on the GSM8K-CoT dataset (0-shot).}
\label{tab:decoupling_ablation}
\vspace{2pt}
\setlength{\tabcolsep}{2.8pt}
\begin{tabular*}{\linewidth}{@{\extracolsep{\fill}}lccc rrr@{}}
\toprule
Method
& \makecell{Bnd.\\aware}
& \makecell{Prefix\\ref.}
& \makecell{Early\\adv.}
& Acc(\%) $\uparrow$
& NFE $\downarrow$
& TPS $\uparrow$ \\
\midrule
Block-wise
& \xmark & \xmark & \xmark
& 73.01 & 337,664 & 41.61 \\
Commit-and-Advance
& \cmark & \xmark & \cmark
& 68.46 & 60,750 & 104.77 \\
\makecell[l]{Boundary-aware\\Coupled}
& \cmark & \cmark & \xmark
& 73.39 & 67,449 & 144.19 \\
\textbf{\method{}}
& \cmark & \cmark & \cmark
& \textbf{74.91} & \textbf{60,378} & \textbf{166.15} \\
\bottomrule
\end{tabular*}
\end{minipage}
\hfill
\begin{minipage}[t]{0.39\linewidth}
\centering
\caption{Asynchrony diagnostics of AsyncLane. We report advancement lead, positive lead ratio, overlap ratio, and fork count on the GSM8K-CoT dataset (0-shot).}
\label{tab:async_diagnostics}
\vspace{2pt}
\setlength{\tabcolsep}{2.8pt}
\begin{tabular*}{\linewidth}{@{\extracolsep{\fill}}lrrrr@{}}
\toprule
Method
& \makecell{Avg.\\lead}
& \makecell{Pos.\\lead}
& Overlap
& \makecell{Fork\\pairs} \\
\midrule
\makecell[l]{Boundary-aware\\Coupled}
& -1.000 & 0.000 & 0.000 & 4,126 \\
Commit-and-Advance
& -- & -- & 0.000 & 0 \\
\textbf{\method{}}
& \textbf{+0.105} & \textbf{0.080} & \textbf{0.074} & 3,906 \\
\bottomrule
\end{tabular*}
\end{minipage}

\end{table}

\paragraph{Boundary Selection Ablation}

\begin{table}[t]
\centering
\scriptsize
\renewcommand{\arraystretch}{1.08}

\begin{minipage}[t]{0.46\linewidth}
\centering
\captionof{table}{
Ablation study on efficient execution mechanisms. We ablate lookahead prefilling and refresh-logit reuse on GSM8K-CoT (0-shot).
}
\label{tab:efficiency_ablation}
\vspace{2pt}
\setlength{\tabcolsep}{3.0pt}

\begin{tabular*}{\linewidth}{@{\extracolsep{\fill}}lrrrr@{}}
\toprule
\textbf{Variant}
& \makecell{\textbf{Acc(\%)}$\uparrow$}
& \makecell{\textbf{Total}\\\textbf{tokens}}
& \makecell{\textbf{NFE/token}$\downarrow$}
& \textbf{TPS} $\uparrow$ \\
\midrule
\textbf{Full \method{}}
& \textbf{74.91}
& \textbf{209,882}
& \textbf{0.288}
& \textbf{166.15} \\

w/o Lookahead
& 67.02
& 123,079
& 0.399
& 117.77 \\

w/o Refresh-Reuse
& 74.37
& 209,613
& 0.312
& 145.94 \\

w/o Both
& 67.02
& 123,079
& 0.424
& 110.58 \\
\bottomrule
\end{tabular*}

\end{minipage}
\hfill
\begin{minipage}[t]{0.52\linewidth}
\centering
\captionof{table}{
Ablation study on boundary selection strategies. We compare period-based, punctuation-augmented, and fixed-interval split rules on GSM8K-CoT (0-shot).
}
\label{tab:boundary_ablation}
\vspace{2pt}
\setlength{\tabcolsep}{2.5pt}

\begin{tabular*}{\linewidth}{@{\extracolsep{\fill}}lrrrrr@{}}
\toprule
\textbf{Boundary rule}
& \makecell{\textbf{Acc(\%)}$\uparrow$}
& $\boldsymbol{\Delta}$ \textbf{Acc}
& \textbf{TPS} $\uparrow$
& \makecell{\textbf{Fork}\\\textbf{count}}
& \makecell{\textbf{Split}\\\textbf{p25/med/p75}} \\
\midrule
Period delimiter
& \textbf{74.91}
& --
& \textbf{166.15}
& 3,906
& 11 / 17 / 24 \\

Punct.-augmented
& 72.48
& $-2.43$ pp
& 149.35
& 8,097
& 11 / 15 / 21 \\

Fixed-$K{=}16$
& 71.34
& $-3.57$ pp
& 147.52
& 8,318
& 16 / 16 / 16 \\
\bottomrule
\end{tabular*}

\end{minipage}

\end{table}

Table~\ref{tab:boundary_ablation} studies how the choice of branch point affects AsyncLane. We compare our period/newline delimiter boundary with two alternatives: a punctuation-augmented and a fixed-position split at $K=16$. Split statistics report the 25th percentile, median, and 75th percentile of fork positions within a lane. The comma and
fixed-position variants produce substantially more fork events, but both reduce accuracy. In particular, comma-based splitting increases the number of forks from 3,906 to 8,097, yet lowers accuracy by 2.43 percentage points.
Fixed-position splitting behaves similarly and yields an even larger accuracy drop. These results indicate that more frequent branching is not sufficient; AsyncLane benefits from branching at low-ambiguity delimiter boundaries that
provide more reliable prefix closure.

\section{Conclusion}
\label{sec:conclusion}

We introduced \method{}, a training-free decoding scheduler for bidirectional diffusion language models. Our key observation is that standard block-wise decoding imposes a block completion barrier: refinement of the current region and advancement of the generation frontier are forced to proceed in a synchronous order. \method{} removes this barrier through active-lane scheduling. When a generate lane exposes a reliable delimiter boundary, branch-and-refine splits the decoding process into a refine lane for the discovered prefix and a
continuation generate lane for the suffix, allowing the future to advance while the prefix remains editable.

To make this asynchronous schedule efficient, \method{} uses shared-prefix lane batching, lookahead draft reuse, cascading termination, and compact cache refresh with refresh-logit reuse. Experiments on mathematical reasoning and code generation benchmarks show that \method{} improves decoding throughput across generation lengths while maintaining competitive quality. These results suggest that relaxing the temporal dependency between refinement and advancement is a promising direction for efficient diffusion language model inference.

A limitation of the current implementation is that boundary selection remains rule-based and task-dependent. While delimiter-based boundaries work well in our experiments, they may not always provide the optimal split point for branch-and-refine. Future work could train a lightweight boundary selector from confidence trajectories or downstream quality signals to decide when and where to fork, potentially improving the speed-quality tradeoff, especially for code
generation and other structurally rich domains.

\bibliographystyle{plainnat}
\bibliography{references}

%%%%%%%%%%%%%%%%%%%%%%%%%%%%%%%%%%%%%%%%%%%%%%%%%%%%%%%%%%%%

% \appendix

% \section{Technical appendices and supplementary material}
% Technical appendices with additional results, figures, graphs, and proofs may be submitted with the paper submission before the full submission deadline (see above). You can upload a ZIP file for videos or code, but do not upload a separate PDF file for the appendix. There is no page limit for the technical appendices. 

% Note: Think of the appendix as ``optional reading'' for reviewers. The paper must be able to stand alone without the appendix; for example, adding critical experiments that support the main claims to an appendix is inappropriate. 

%%%%%%%%%%%%%%%%%%%%%%%%%%%%%%%%%%%%%%%%%%%%%%%%%%%%%%%%%%%%

% \newpage
% \input{checklist.tex}

\end{document}